\documentclass{article} 
\usepackage{iclr2026_conference,times}

\usepackage{amsmath,amsfonts,bm}

\def\eqref#1{equation~\ref{#1}}

\def\1{\bm{1}}

\DeclareMathAlphabet{\mathsfit}{\encodingdefault}{\sfdefault}{m}{sl}
\SetMathAlphabet{\mathsfit}{bold}{\encodingdefault}{\sfdefault}{bx}{n}

\usepackage{hyperref}
\usepackage{url}

\usepackage{array}
\usepackage{booktabs}
\usepackage{graphicx}
\usepackage{subcaption}
\usepackage{cleveref}
\newcommand*\chem[1]{\ensuremath{\mathrm{#1}}}

\usepackage{xcolor}
\usepackage{tikz}
\usetikzlibrary{positioning,calc,arrows.meta,fit,backgrounds}

\definecolor{lfmWAC}{RGB}{108,100,128}
\definecolor{lfmLOLA}{RGB}{80,128,95}
\definecolor{lfmDiviner}{RGB}{60,111,143}
\definecolor{lfmMiniRF}{RGB}{143,103,192}
\definecolor{lfmGRAIL}{RGB}{186,115,95}
\definecolor{lfmClem}{RGB}{154,130,95}

\definecolor{lfmBlueLight}{RGB}{243,247,255}
\definecolor{lfmYellowLight}{RGB}{255,248,235}
\definecolor{lfmGreenLight}{RGB}{238,247,240}
\definecolor{lfmRedLight}{RGB}{255,242,241}
\definecolor{lfmPanel}{RGB}{251,252,254}
\definecolor{lfmGridGreen}{RGB}{155,209,139}
\definecolor{lfmGridYellow}{RGB}{241,217,111}
\definecolor{lfmGridOrange}{RGB}{244,162,97}

\title{LunarFM: A Shared Multimodal Representation of the Moon's Surface}

\author{Marc Girona-Mata\thanks{Equal contribution} \\
University of Cambridge \\
\And
$\text{Jakob Gawlikowski}^{*}$\\
German Aerospace Center (DLR) \\
\And
Sumit Goski\\
SPAIDER SPACE \\
\And
Gautier Bardi de Fourtou\\
Mines Paris - PSL University \\
\And
Valentin T. Bickel\\
University of Bern \\
\And
Ben Moseley\\
Imperial College London \\
\AND
Abigail Calzada-Diaz  \\
European Space Resources \\
Innovation Center (ESRIC) \\
\And
Sylvester Kaczmarek \\
Imperial College London \\
\And
Raúl Ramos-Pollán\thanks{Corresponding author} \\
Universidad de Antioquia \\ 
$\texttt{raul.ramos@udea.edu.co}$
}
\iclrfinalcopy 
\begin{document}

\maketitle
\begin{abstract}
The renewed global focus on lunar exploration, driven by the prospect of in-situ resource utilization and a sustained human presence on the Moon, has created growing demand for accurate, large-scale characterization of the lunar surface. Although vast quantities of orbital remote-sensing data have been collected, scientific analysis and resource mapping remain fragmented by heterogeneous multi-instrument observations, sparse labels, and bespoke task-specific modelling workflows. Here we introduce LunarFM\footnote{An earlier version of this work appeared as a workshop paper in the ICLR 2026 Workshop FM4Science. This revised version includes enhanced figures and discussion, as well as an additional experiment showcasing a geological classification downstream task.}, a multimodal foundation model that learns a general representation of the lunar surface from diverse orbital measurements. LunarFM assimilates observations from six instruments across three lunar missions, mapping 18 input channels to a shared embedding space. We demonstrate that this embedding space supports a diverse range of downstream applications, including similarity search, few-shot resource mapping, mineral abundance regression, and geological unit classification, enabling efficient scientific investigation and resource-oriented analysis. We provide a machine-learning-ready dataset of co-registered multimodal observations spanning latitudes from 70°S to 70°N, a pretrained multimodal masked autoencoder, and a companion embedding dataset providing a joint 768-dimensional representation of lunar surface properties. All code and data are available at \url{https://lunarfm.trillium.tech/}

\end{abstract}

\section{Introduction}
Over the last two decades, the number of missions to the Moon has skyrocketed. Besides scientific interests, this renewed activity is primarily driven by two goals: exploring the Moon as a potential source of valuable resources and establishing a long-term, sustained lunar station as a base for deeper space exploration.
In 2009, the Indian mission Chandrayaan-1 detected signatures of \chem{H_2O} and \chem{OH} in the southern circumpolar regolith \citep{Pieters2009, Colaprete2010}; this was confirmed a little later by NASA’s Lunar Crater Observation and Sensing Satellite (LCROSS) mission, which unambiguously detected water ice in Cabeus Crater close to the lunar south pole \citep{neish2011nature}.
Besides water ice, the Moon is known to contain many other useful materials. Ilmenite (titanium dioxide, \chem{TiO_2}), silicon, and aluminium (e.g., \chem{Al_2O_3}) are abundant in lunar regolith and can be used for construction \citep[e.g., solar panels;][]{cuervo2025moon} and life-support applications \citep[e.g., oxygen, rocket fuel][]{sargeant2020hydrogen}. Other elements, such as rare-earth metals \citep{hedrick2023reemoon} and helium-3, could be used in advanced technologies and energy production.

Lunar geology and resource maps are typically derived from orbital remote-sensing observations collected by multiple missions, including spectral, gamma-ray, neutron, and topographic measurements, among others \citep[e.g.,][]{sato2017tio2}. From these primary observations, a range of secondary and higher-level products are generated through task-specific processing pipelines. In practice, however, most of these output products rely on a limited combination of datasets, rather than fully exploiting the large amount of multi-modal information available. 
This is not due to a lack of available data, but rather to the challenges of integrating heterogeneous observations. Lunar datasets differ substantially in spatial and temporal resolution, observation geometry, radiometric characteristics, signal-to-noise properties, processing assumptions, and spatial-temporal surface coverage. Combining these datasets in a physically-consistent and statistically-rigorous manner therefore requires significant expertise and task-specific engineering. Furthermore, when generating output products, high-quality labels remain scarce, necessitating considerable effort to translate them into output products. As a result, existing scientific analysis and resource mapping workflows remain fragmented and inefficient, each requiring bespoke data integration and modelling workflows.

To this end, we introduce LunarFM, a multimodal foundation model that learns a general representation of the lunar surface from diverse orbital measurements. LunarFM assimilates observations from six instruments across three lunar missions, yielding a total of 18 input channels/bands, and maps them to a shared embedding space that can be used to support a diverse range of downstream applications, including similarity search, few-shot resource mapping, mineral abundance regression, and geological unit classification. Our framework substantially reduces the need for task-specific workflow development and enables efficient scientific investigation and resource-oriented analysis.

Our contributions are:
\begin{enumerate}
    \item a curated, integrated, and machine learning-ready dataset (LunarChips) for LunarFM pretraining that includes data from three lunar orbital remote-sensing missions and six instruments/modalities, spanning latitudes from 70°S to 70°N and divided into chips that extend 0.5 degrees of latitude and longitude each, along with additional datasets for evaluation;
    \item a pretrained multimodal masked autoencoder and companion embedding dataset (LunarEmbeddings), providing 768-dimensional dense representations for each (18-channel) 0.5° input chip; and
    \item a set of evaluation experiments on illustrative downstream applications that demonstrate the potential of the learned representations, including multimodal input reconstruction, unsupervised exploration of embedding representations, similarity search, mineral mapping, and geological classification.
\end{enumerate}

Alongside these resources, we provide tutorials for hands-on application of these resources to downstream tasks.

In the following, we introduce the data sources and preparation used for training LunarFM, describe the model's multimodal architecture and training strategy,  explore the resulting embedding representations, and demonstrate illustrative downstream applications. We discuss design choices and current limitations throughout, and frame open questions for future work in Section~\ref{sec:discussion}.

\section{Related Work}
Foundation models (FMs) are large-scale, pretrained models that have demonstrated remarkable capabilities across diverse domains \citep{wang_self-supervised_2022, tao_self-supervised_2023}. Their rise has also been seen in Earth observation (EO), with recent developments showing promising results \citep{jakubik2025terramind, astruc2025anysat, tsenggalileo}.
Building on these advances, geospatial foundation models (GFMs) such as CROMA \citep{fuller2023cromaremotesensingrepresentations}, DOFA \citep{xiong2025neuralplasticityinspiredmultimodalfoundation}, TerraMind \citep{jakubik2025terramind}, AnySat \citep{astruc2025anysat}, and Galileo \citep{tsenggalileo} have emerged, integrating multiple data sources such as optical, radar, digital elevation models (DEMs), land-cover, and text, into unified architectures capable of sensor-agnostic understanding across diverse downstream tasks.

This trend is beginning to extend beyond EO into planetary science. For Mars, \citet{fang2026domain} develop a domain-specific vision foundation model trained with self-supervised learning on large-scale Martian imagery, showing that planetary-domain pretraining can outperform generic visual pretraining for Mars surface analysis. Also for Mars, \citet{purohit2026momo} introduce MOMO, a multi-sensor foundation model that combines representations from multiple orbital instruments to support a range of downstream remote-sensing tasks. These works provide early evidence that foundation-model ideas can transfer effectively to planetary remote sensing, while also highlighting the importance of domain-specific data and multi-instrument learning.

For multimodal representation learning across lunar instruments, \cite{sander2025moonsfacessingleunified} trained a transformer model on grayscale images, surface normals, DEMs, and albedo maps, with a focus on reconstructing DEMs and albedo from reflectance imagery. LunarFM extends this direction by integrating a broader set of physically distinct modalities, spanning optical, thermal, radar, topographic, and gravitational data, within a shared representation space designed to support a range of downstream tasks. We note that our work is complementary to \cite{prasad2026moonstone}, who recently released a similar effort in terms of training a multimodal embeddings model of the Moon, alongside a set of benchmark downstream tasks. 

Lastly, an adjacent line of work is emerging around planetary vision-language models. In the lunar domain, \citet{inal2026llava} introduce LLaVA-LE, a large language-and-vision assistant for lunar exploration, together with LUCID, a multimodal dataset of high-resolution lunar images, captions, and question-answer pairs for domain-specific visual instruction tuning.

\section{Methods} \label{sec:technical-setup}

\subsection{Overview} 

\begin{figure}[h!]
    \centering
    \includegraphics[width=\textwidth]{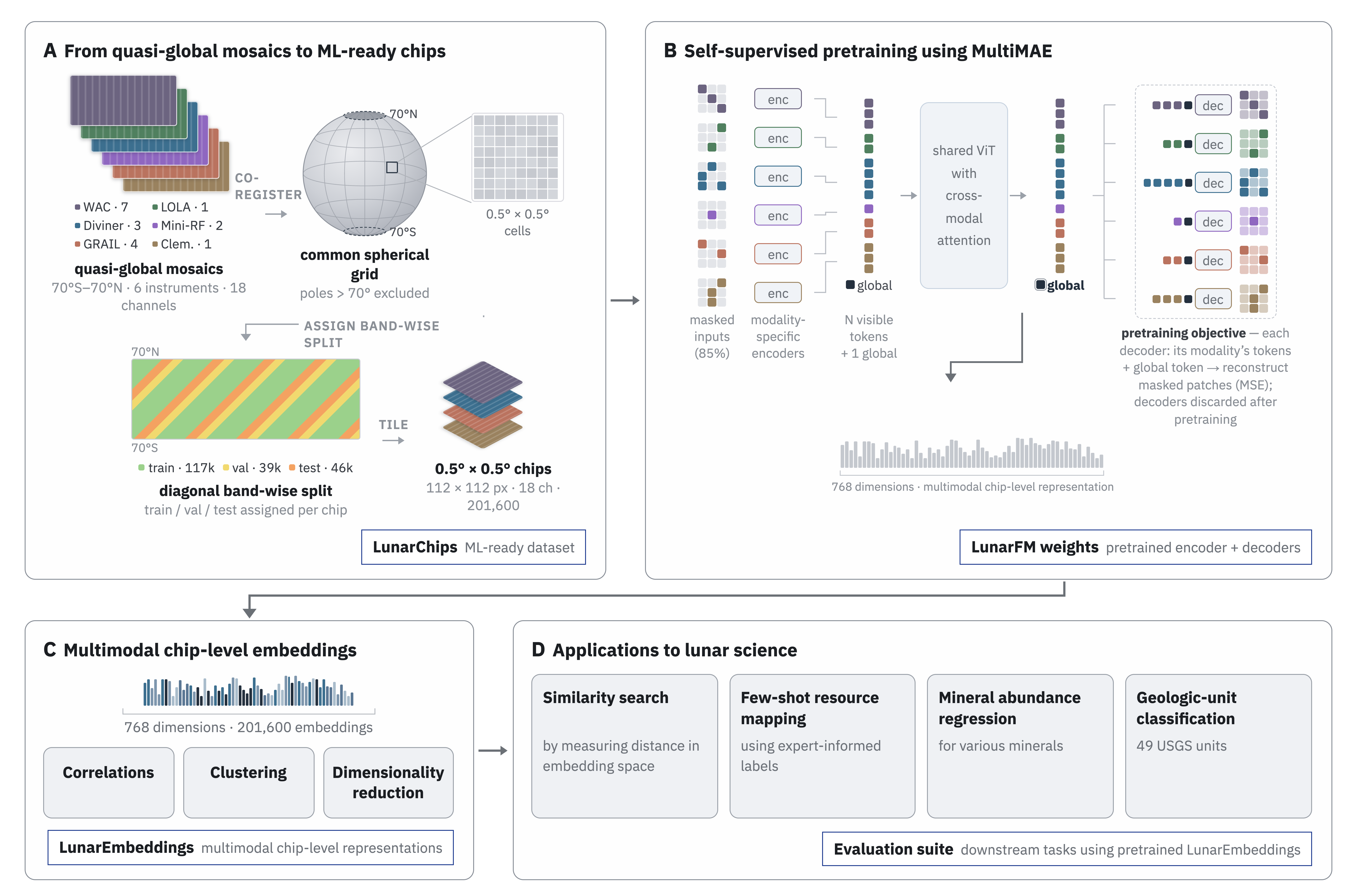}
    \caption{Overview of the LunarFM ecosystem. (A) Quasi-global mosaics (70°S–70°N) from six instruments across three missions (18 channels total) are co-registered onto a common spherical grid, split diagonally into train/validation/test bands, and tiled into 0.5°×0.5° chips; released as the LunarChips dataset. (B) A MultiMAE-style model is pretrained via self-supervised learning: masked, modality-specific inputs are encoded, fused by a shared ViT processor into patch-level and global tokens, and reconstructed per modality (with decoders discarded after pretraining), yielding the pretrained LunarFM weights. (C) Encoding every chip with the pretrained LunarFM weights produces one 768-dimensional global embedding per chip (the LunarEmbeddings), a multimodal, chip-level representation amenable to correlation analysis, clustering, and dimensionality reduction. (D) Lightweight models trained on the frozen embeddings (no fine-tuning) support downstream lunar-science tasks: similarity search, few-shot resource mapping, mineral-abundance regression, and geologic-unit classification.}
    \label{fig:lunarfm-overview}
\end{figure}

\Cref{fig:lunarfm-overview} summarises the LunarFM workflow. Multimodal observations from six instruments are first assembled into the LunarChips dataset, which is used to self-supervise a multimodal masked autoencoder. The pretrained encoder then produces a 768-dimensional embedding for every lunar chip, forming the LunarEmbeddings dataset. These embeddings provide a compact, general-purpose representation of the lunar surface that can be used directly by lightweight models for a variety of downstream lunar science tasks.

\begin{table}[t]
    \centering
    \caption{Overview of selected orbital remote-sensing instruments and data modalities integrated into the LunarFM model. Note that Diviner Rock Abundance (RA) is treated as a separate input group during training, yielding seven input groups in total.}
    \label{tab:lunar_instruments}
    \resizebox{\textwidth}{!}{
    \begin{tabular}{>{\raggedright\arraybackslash}p{2.5cm} >{\raggedright\arraybackslash}p{2cm} >{\raggedright\arraybackslash}p{2cm}>{\raggedright\arraybackslash}p{1.5cm} >{\raggedright\arraybackslash}p{5cm}}
\toprule
\textbf{Instrument} & \textbf{Mission} & \textbf{Modality} & \textbf{Channels} & \textbf{Description} \\
\midrule

LROC WAC & LRO & Multispectral imagery &7 & Photometrically Normalised Mosaic via Hapke Parameter Maps \\[15pt]

LOLA & LRO &Topography elevation & 1 & Surface digital elevation model (DEM) \\[15pt]

Diviner & LRO & Thermal emission & 3 & Regolith temperature, rock abundance, bolometric temperature \\[15pt]

Mini-RF &LRO & Radar & 2 & Circular polarization ratio and S1 (first Stokes parameter) \\[15pt]

Ka-band Lunar Gravity Ranging System &GRAIL & Gravitational anomaly & 4 & Free-air gravity anomaly, Bouguer anomaly, disturbance, and measurement uncertainty \\[30pt]

UVVIS Camera &Clementine & Albedo & 1 & Brightness measured at the 750 nanometer (nm) wavelength \\[15pt]

\bottomrule
\end{tabular}}
\end{table}
\begin{figure}[h!]
    \centering
    \begin{subfigure}[b]{0.32\textwidth}
        \scalebox{1}[-1]{\includegraphics[width=\textwidth, height=2.5cm]{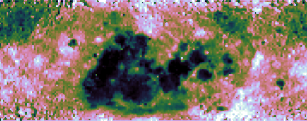}}
        \subcaption{LROC WAC (visible)}
    \end{subfigure}
    \begin{subfigure}[b]{0.32\textwidth}
        \scalebox{1}[-1]{\includegraphics[width=\textwidth, height=2.5cm]{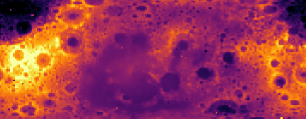}}
        \subcaption{LOLA (topographic elevation)}
    \end{subfigure}
    \begin{subfigure}[b]{0.32\textwidth}
        \scalebox{1}[-1]{\includegraphics[width=\textwidth, height=2.5cm]{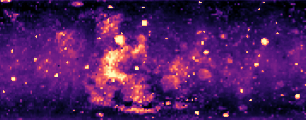}}
        \subcaption{Diviner (rock abundance)}
    \end{subfigure}
    \begin{subfigure}[b]{0.32\textwidth}
        \scalebox{1}[-1]{\includegraphics[width=\textwidth, height=2.5cm]{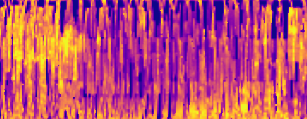}}
        \subcaption{Mini-RF (S1)}
    \end{subfigure}
    \begin{subfigure}[b]{0.32\textwidth}
        \scalebox{1}[-1]{\includegraphics[width=\textwidth, height=2.5cm]{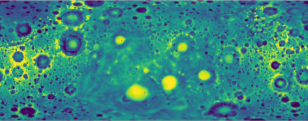}}
        \subcaption{GRAIL (inv. density)}
    \end{subfigure}
    \begin{subfigure}[b]{0.32\textwidth}
        \scalebox{1}[-1]{\includegraphics[width=\textwidth, height=2.5cm]{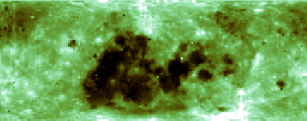}}
        \subcaption{Clementine (visible)}
    \end{subfigure}
    \caption{Maps showing single-channel examples for the six instruments included in the LunarChips dataset and used for the pretraining LunarFM, covering most of the Moon's surface, from -70° to +70° latitude. For LunarFM,  Diviner Rock Abundance is treated as an individual seventh input.
}
    \label{fig:data_examples}
\end{figure}
\subsection{Dataset Preparation}

For training LunarFM, we combine data from six modalities across three missions: the Lunar Reconnaissance Orbiter (LRO), the Gravity Recovery and Interior Laboratory (GRAIL), and the Clementine missions. An overview of the considered missions and sensing instruments is provided in Table \ref{tab:lunar_instruments}, and selected examples are shown in \Cref{fig:data_examples}. Further details are included in Appendix \ref{app:data_sources}.

Data was obtained from NASA's Planetary Data System (PDS) Geoscience Node\footnote{https://pds-geosciences.wustl.edu/dataserv/moon.html}, either from global mosaics or other derived products as available at PDS. We extract 0.5° $\times$ 0.5° longitude-latitude chips from the data using a spherical grid discretisation of the Moon's surface and store them as individual GeoTIFF files. For training and inference, each channel is normalised independently using band-wise mean and standard deviation computed across the training set. Due to data availability, resolution, and distinct solar illumination effects, we focus on the region between -70° and +70° latitude, resulting in a total of 201,600 chips. We further apply a band-wise split to separate grid cells into training, validation, and test sets of 117,024, 38,715, and 45,861 samples, respectively (see Appendix \ref{app-sec:data-split}). %
The band-wise split is used to mitigate spatial leakage, thereby enabling robust evaluation of the model's generalisability and reducing overfitting. The resulting data is part of our LunarChips dataset. 

Additionally, LunarChips includes products for downstream applications. These include global elemental and compositional maps derived from gamma-ray spectroscopy \citep{prettyman2006_elemental_composition_lunar_prospector}, WAC-derived ilmenite estimates, impact crater annotations, and chip-aligned geological-unit labels derived from USGS lunar geological mapping products. We further incorporate the delineations of the high-ilmenite (\chem{TiO_2}) region proposed by \cite{diaz2025descriptive}, which we treat as an expert-curated reference set for low-label resource mapping experiments. Collectively, these layers enable consistent, chip-aligned supervision and evaluation across tasks, including mineral regression, few-shot resource screening, crater-related analyses, and geological-unit classification, while keeping the representation-learning stage strictly self-supervised for the six primary sensing modalities.

\subsection{Model Architecture and Training}
To compute the joint representation, we use a multimodal multi-masked autoencoder \citep[MultiMAE;][]{bachmann2022multimae} with visual transformer backbones as the encoder and decoder. The input chips are grouped into seven groups (with each modality grouping channels corresponding to the same instrument, except for Diviner, which is split into two modalities) and divided into patches of $8\times 8$ pixels, followed by a modality-wise linear projection of each patch to an input token with dimension 768. Linear projections are optimised during training individually for each input data source. Spatial information is added using sinusoidal positional embeddings, which encode the patch's relative position within the chip. Data source information is preserved by modality-wise linear mappings that learn to encode modality-specific bias values. In addition to the patch-level tokens in the input data, we extend the input representation with a global token that is independent of modality or patch location. In the embedding space, this global token serves as the chip-level embedding. Following \cite{bachmann2022multimae}, the input tokens are processed by a joint encoder network based on a visual transformer, which yields latent embedding vectors that capture both single-modality and cross-modality information. Given the available embeddings and the global (modality-independent) token, lightweight modality-wise decoders aim to reconstruct the original input. The final model consists of 109.8 million parameters.

During training, we augment the training data by first slicing the Moon into 2° chips and then randomly selecting 0.5° footprints within each 2° chip. Missing values are set to zero and excluded from the reconstruction loss, such that NaN regions do not contribute to the gradient during training. Chips are aligned and resized modality-wise to $112 \times 112$ pixels. The patch size is set to $8 \times 8$ pixels, yielding 196 patches ($14 \times 14$) per 0.5° input chip. During training, we randomly mask 85\% of the input patches across all input modalities, while the decoder reconstructs the full input domain, including masked locations, thereby favouring the learning of intra- and cross-modality correlations. Training minimises the mean squared reconstruction error, weighted equally across modalities, using the Adam optimiser with an initial learning rate of $10^{-4}$, reduced every 50,000 iterations. The model is trained for approximately 500,000 iterations with a batch size of 32 per GPU on three NVIDIA H100 GPUs.

\section{Evaluation}

\subsection{Reconstruction Error}
\begin{figure}[h!]
    \centering
    \includegraphics[width=\linewidth,trim="0cm 2cm 0cm 0cm", clip]{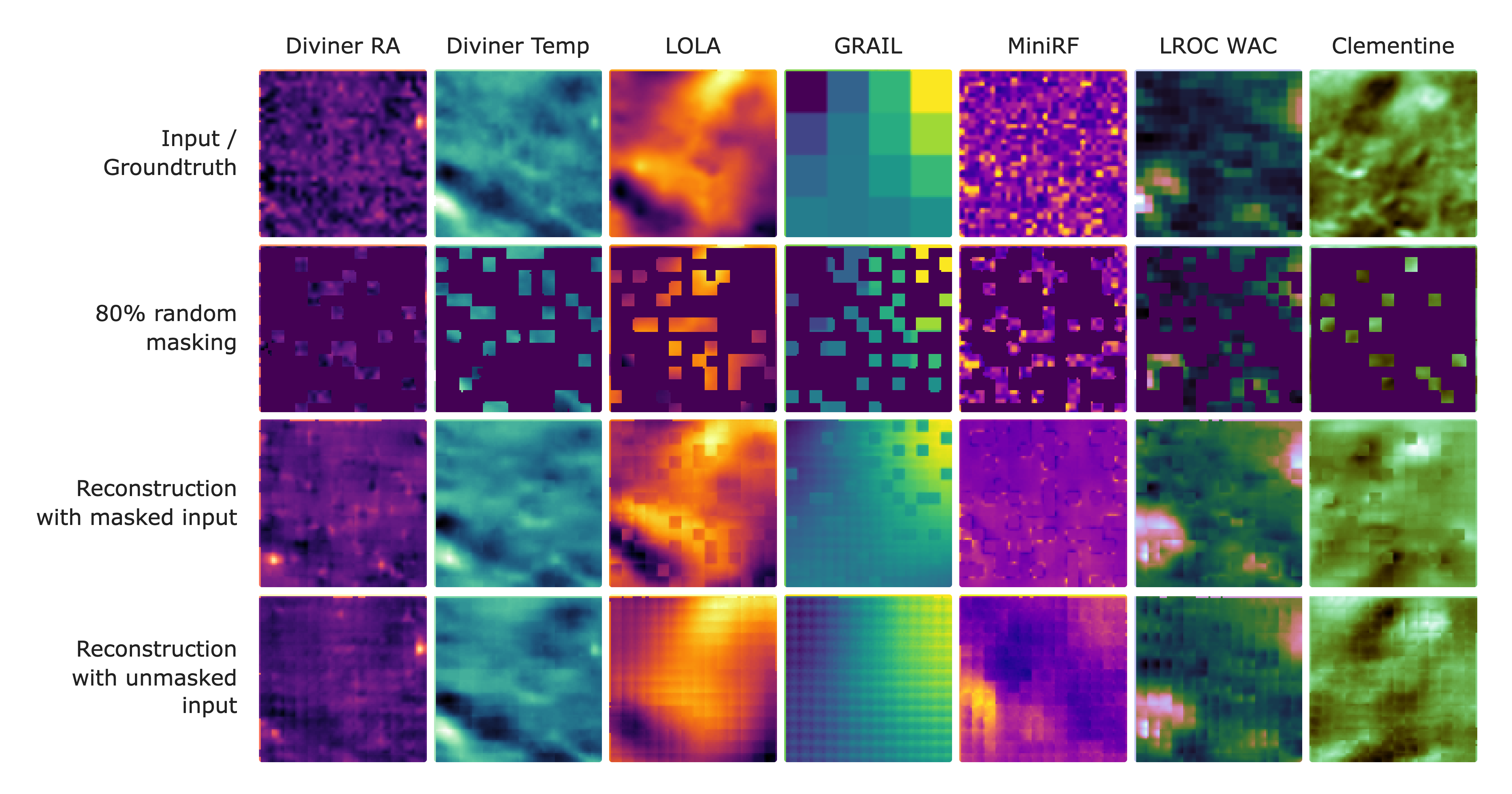}

    \caption{An example of modality-wise reconstruction of a 112 by 112 pixel chip with 80\% masking (third row) and without masking (fourth row). Note that the model is trained with 85\% masking.}
    \label{fig:reconstruction_example_masked}
\end{figure}

First, we evaluate the reconstruction performance as a proxy for the quality of the learned representations, noting that perfect reconstruction is not the model's primary goal. \Cref{tab:reconstruction-error} shows the average squared error in the reconstruction for the individual modalities, both when the modality is present and when it is withheld. As expected, the error is higher when reconstructing a missing modality. For Mini-RF, the error is higher than for the others, which can be explained by the nature of Mini-RF and various NaN values in the data (see also \Cref{fig:data_examples} (d)). 
\Cref{fig:reconstruction_example_masked} shows an example for a reconstruction with unmasked and 80\% masked-out inputs, underlining the learned correlations among the modalities.   

\begin{table}[htbp]
  \centering
  \caption{Modality-wise reconstruction error using all modalities as input (top row) and excluding the reconstructed modality from the input (bottom row). The reconstruction error is expressed as mean squared error with respect to standardised inputs with zero mean and unit variance, based on the training data statistics. NaN input pixels are excluded from the computation.}
  \label{tab:reconstruction-error}
  \setlength{\tabcolsep}{4pt}
  \renewcommand{\arraystretch}{1.2}
  \resizebox{\textwidth}{!}{%
  \begin{tabular}{l*{7}{c}}
    \toprule
     & \shortstack{Diviner\\RA} & \shortstack{Diviner\\Temp} & LOLA & GRAIL & MiniRF & \shortstack{LROC\\WAC} & Clementine \\
    \midrule
    With modality    & \shortstack{$0.092\ \pm$\\$0.085$} & \shortstack{$0.008\ \pm$\\$0.004$} & \shortstack{$0.001\ \pm$\\$0.001$} & \shortstack{$0.002\ \pm$\\$0.003$} & \shortstack{$0.420\ \pm$\\$0.190$} & \shortstack{$0.026\ \pm$\\$0.015$} & \shortstack{$0.022\ \pm$\\$0.015$} \\
    \addlinespace
    Without modality & \shortstack{$0.203\ \pm$\\$0.171$} & \shortstack{$0.055\ \pm$\\$0.035$} & \shortstack{$0.007\ \pm$\\$0.008$} & \shortstack{$0.266\ \pm$\\$0.304$} & \shortstack{$0.440\ \pm$\\$0.210$} & \shortstack{$0.059\ \pm$\\$0.050$} & \shortstack{$0.067\ \pm$\\$0.069$} \\
    \bottomrule
  \end{tabular}%
  }
\end{table}

\subsection{Lunar Embeddings}

\begin{figure}[h!]
    \centering
    \includegraphics[width=\textwidth]{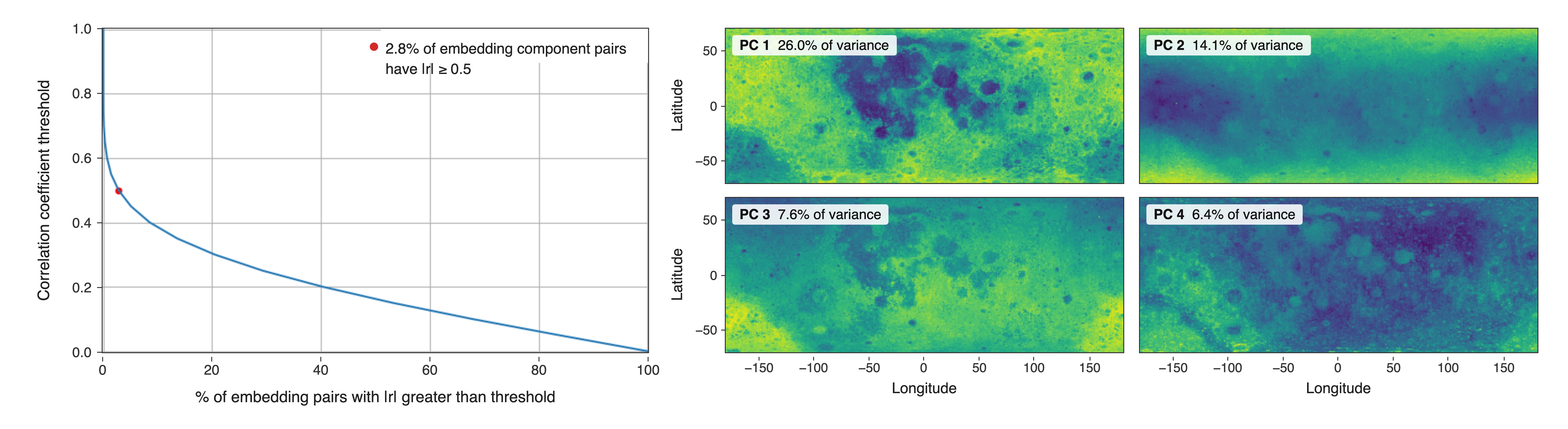}
    \caption{Analysis of the internal structure of the embeddings. \textbf{Left:} the distribution of correlation coefficients, highlighting a rapid decay and low redundancy between dimensions.
    \textbf{Right:} Maps of the first four principal components (PCs), which cumulatively explain 54.1\% of the variance. The first four principal components appear to encode topographic elevation and surface albedo, surface temperature and effective albedo, surface roughness and gravity, and gravity, respectively.}
    \label{fig:cross_correlation_and_pca}
\end{figure}

\paragraph{Correlation.}
To assess first insights into the information density of the 768-dimensional feature space, we computed the Pearson correlation coefficient ($r$) between all pairs of embedding dimensions. \Cref{fig:cross_correlation_and_pca} (left) displays the distribution of these coefficients, which decays rapidly, with only 2.8\% of embedding component pairs exhibiting an absolute correlation $|r| \ge 0.5$. This low degree of cross-correlation indicates that the learned embedding dimensions are largely orthogonal, suggesting the model avoids trivial linear redundancy, however we note that low pairwise (linear) correlation is a necessary rather than sufficient condition for information richness.

\paragraph{Dimensionality reduction.}
We use two distinct dimensionality reduction techniques, namely Principal Component Analysis (PCA) and Uniform Manifold Approximation and Projection (UMAP), to visualise the high-dimensional embedding space with the intent to gain an intuitive understanding of how the embeddings are able to capture geological features on the Moon. We employ this combination to capture different aspects of the data topology: PCA reveals global linear variations, whereas UMAP exploits non-linearity to highlight both local structure whilst retaining a global continuous manifold.
We extracted the first four principal components (PCs) to analyse the dominant linear variations in the data. Together, these four components account for approximately 54.1\% of the total variance in the dataset and are visualised in \Cref{fig:cross_correlation_and_pca} (right).
The first component alone accounts for 26.0\% of the variance, with subsequent components (PC2 through PC4) explaining 14.1\%, 7.6\%, and 6.4\%, respectively. The spatial distribution (and coherence) of these components suggests that the orthogonal directions of maximum variance in the embedding space correspond to large-scale lunar surface features. These four components appear to be influenced by distinct input data, namely surface albedo and elevation (PC1), effective albedo and surface temperature (PC2), gravity and roughness (PC3), and a strong gravity component (PC4).

\begin{figure}[h!]
    \centering
    \includegraphics[width=\textwidth]{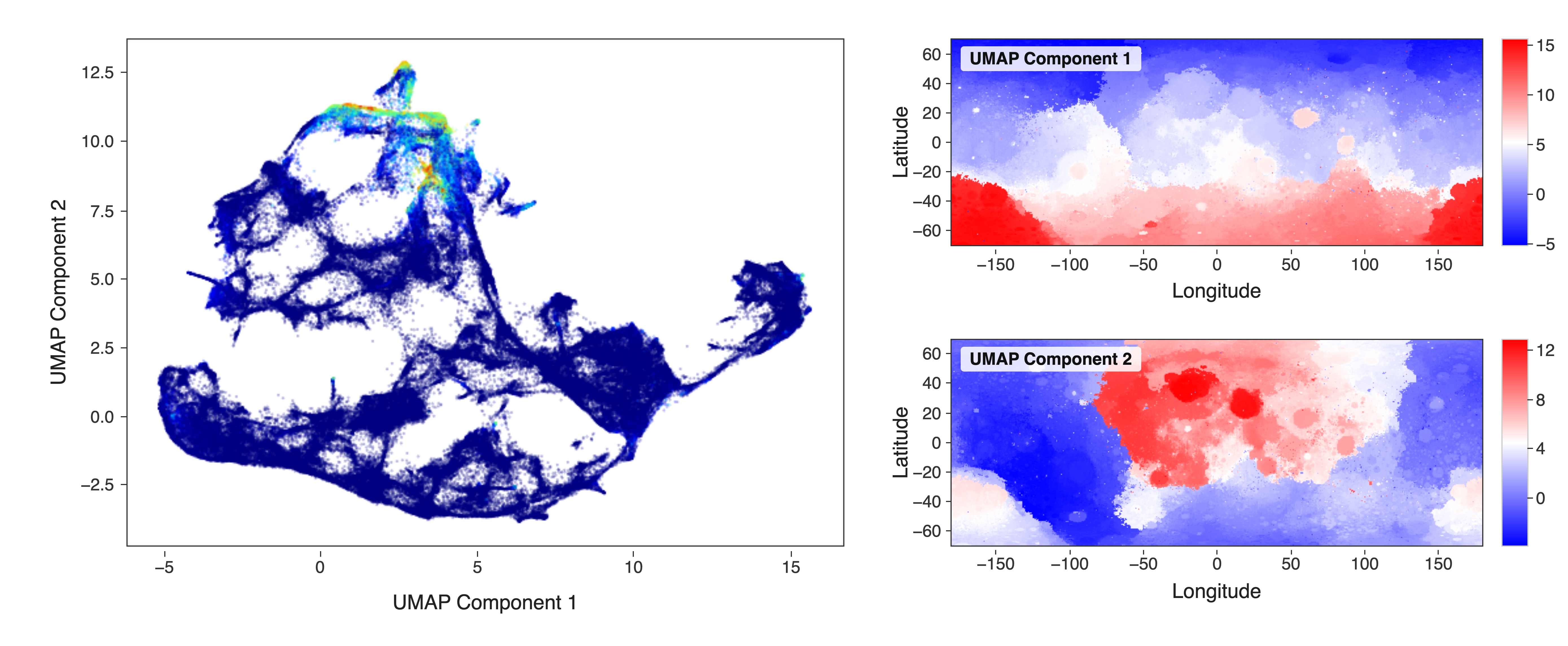}
    \caption{UMAP analysis of lunar embeddings. Left: 2D scatter plot of the projected embeddings, coloured by WAC-derived ilmenite (\chem{TiO_2}) percentage. Observe how chips with high \chem{TiO_2} are naturally separated in this projection, whilst nowhere in the training process \chem{TiO_2} information was shown to the model. Right: Spatial distribution of the first and second UMAP components mapped to the lunar grid. Observe how major geological features of the Moon naturally arise, such as the maria and the South Pole Aitken (SPA) Basin.}
    \label{fig:umap_composite}
\end{figure}

To investigate non-linear relationships, we project the embeddings into two dimensions using UMAP. 
The two components visualised in \Cref{fig:umap_composite} (left), show a continuous manifold structure. The spatial distribution of the two UMAP components aligns with surface features, indicating that the embeddings encode smooth transitions among lunar terrains while maintaining distinct boundaries between disparate geological zones. \Cref{fig:umap_composite} also shows the colouring of the space by ilmenite (\chem{TiO_2}) concentration, revealing a small, relevant subregion in this distribution. These structures also show differentiated grouping when using other reference sets for different minerals. When these two UMAP dimensions are mapped back onto the lunar surface (right part of \Cref{fig:umap_composite}), they exhibit strong spatial contiguousness, differentiating major geological regions despite the model having no explicit coordinate input during inference. Overall, this provides evidence that the embeddings are sensitive to different physical properties.

\begin{figure}[h!]
    \centering
    \includegraphics[width=\textwidth]{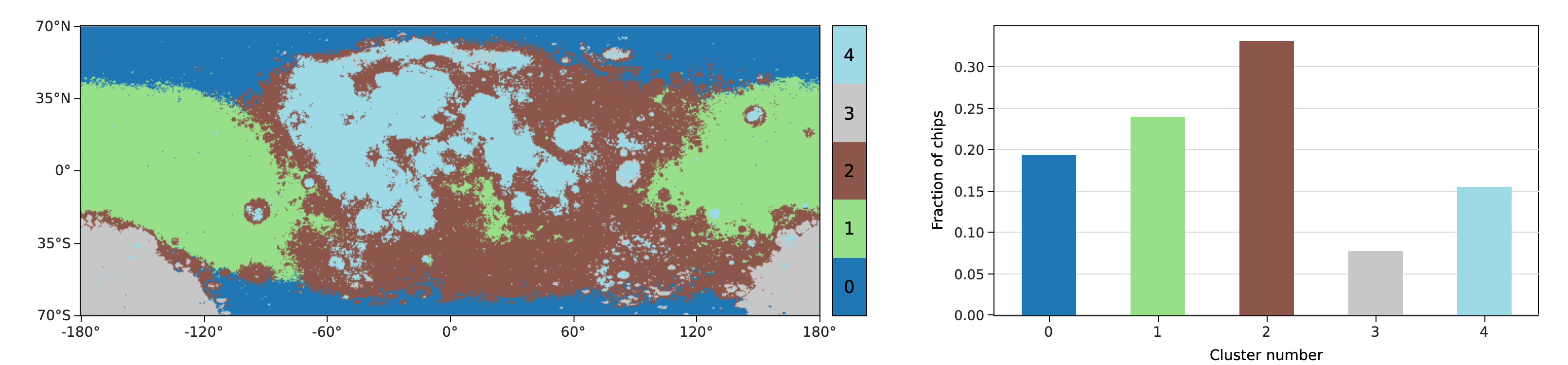}
    \caption{Unsupervised K-Means clustering ($k=5$) of the lunar embeddings, revealing five distinct clusters: 0) circumpolar regions, 1) equatorial farside, 2) nearside highlands, 3) South Pole-Aitken Basin, and 4) mare regions. Left: The spatial distribution of the assigned clusters. Right: The fraction of total lunar chips assigned to each cluster.}
    \label{fig:kmeans_clustering}
\end{figure}

\paragraph{Clustering.}
Finally, we apply k-means clustering ($k=5$) directly to the 768-dimensional embeddings. \Cref{fig:kmeans_clustering} illustrates the resulting cluster map and the distribution of chips across clusters. The resulting clusters map to spatially-coherent (and geographically meaningful) regions without any neighbourhood constraints enforced during the clustering process. The distribution of chips shows that while some clusters represent vast, dominant terrain types (e.g., cluster 2), others capture rarer surface characteristics (e.g., cluster 3), suggesting that the model can discriminate between common and unique lunar features.

\subsection{Example Downstream Applications}\label{sec:downstream}
In this section, we present four illustrative downstream applications that demonstrate how LunarFM embeddings can serve as a starting point for resource-oriented analysis.

\subsubsection{Similarity Search}
\begin{figure}[h!]
    \centering
    \includegraphics[width=\textwidth]{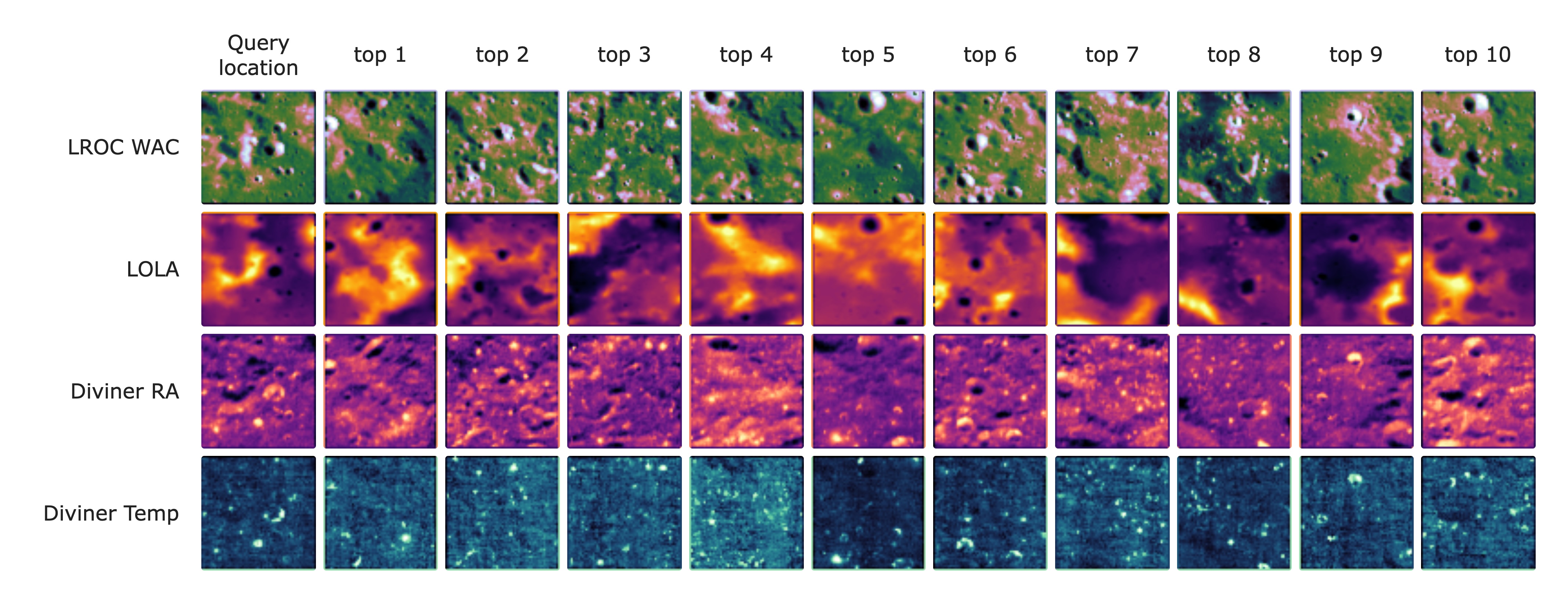}
    \caption{An example of a similarity search, four modalities of the query location are shown together with the top ten matches. The rows from top to bottom show LRO WAC, LOLA DEM, Diviner Regolith Temperature, and Diviner Rock Abundance. }
    \label{fig:similarty_search}
\end{figure}

First, latent spaces of large models trained via self-supervised learning have been shown to provide good premises for an efficient similarity search. If images share information in their inputs, then this information should also lead to a corresponding similarity in the embeddings that encode these features. In \Cref{fig:similarty_search} an example for similarity search based on the L2 distance among the z-score normalised embeddings is shown. For the moment, we keep this assessment qualitative as shown in the figure, whilst a more formal retrieval metric on the Moon might be developed in the future.

\subsubsection{Mineral composition}

\begin{figure}[h!]
    \centering
    \includegraphics[width=\textwidth]{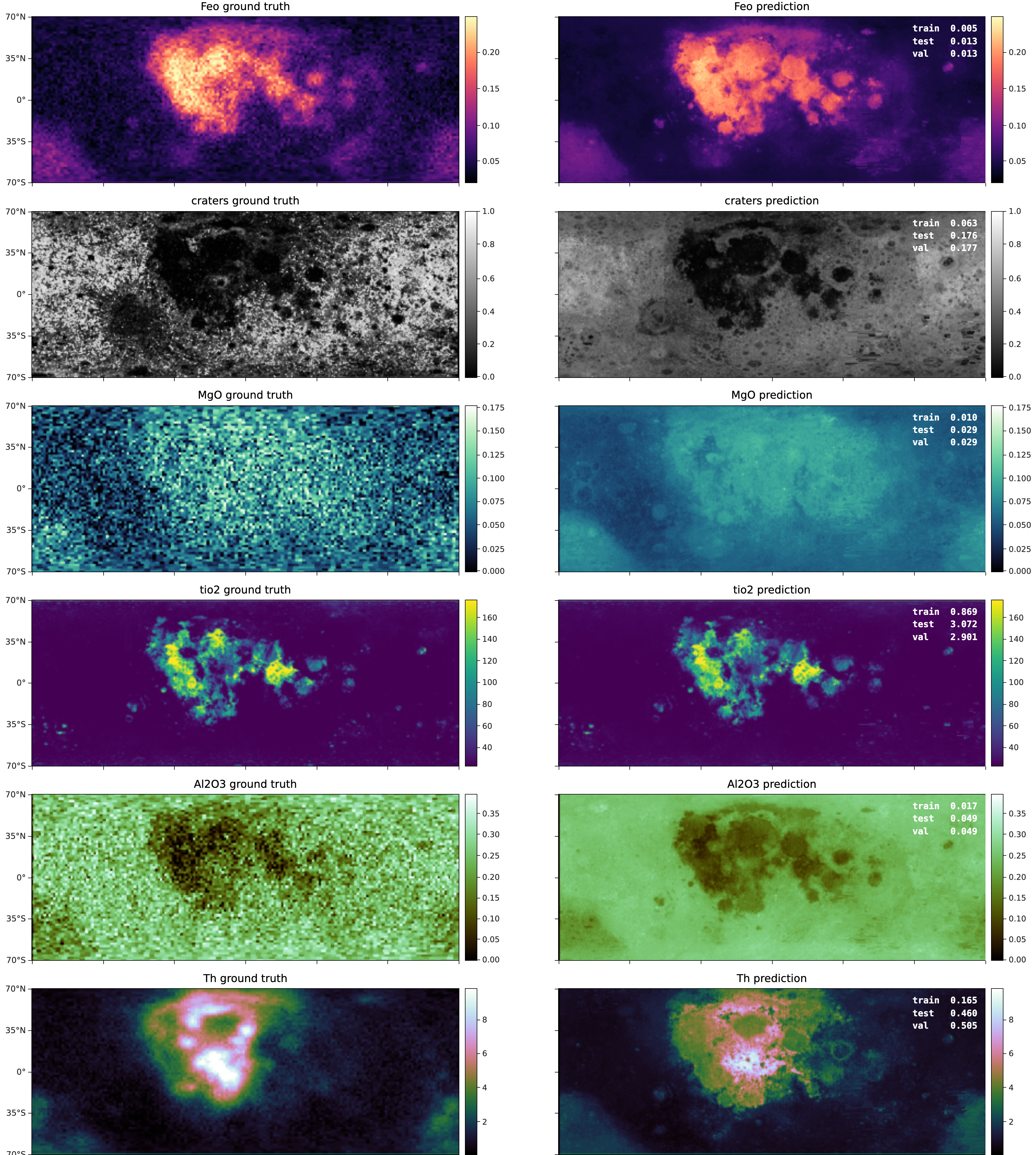}
    \caption{Results on different downstream tasks predicting the presence of different minerals on the Moon's surface, based on random forest. In the upper right of the predicted maps, the Mean Absolute Errors of the predictions are stated. Predictions are produced on the finer LunarChips grid, although this finer output sampling should not be interpreted as an increase in the effective spatial resolution of the reference data.}
    \label{fig:downstream_tasks}
\end{figure}

The second task is mineral composition mapping. The experimental setting is as follows. To begin with, we obtain embeddings for all lunar chips. Then, we randomly select a subset of 20,000 chips from the training split. Subsequently, we train a simple regression model (random forest) using the embeddings as inputs to predict the average mineral content per chip on the training data subset. Finally, we use the trained random forest model to predict mineral content for the remaining chips. \Cref{fig:downstream_tasks} shows the different mineral compositions we used to test this approach. Again, our intent is to show that a simple regression model on our embeddings can produce meaningful mineral distribution maps. Future work could be targeted to specific applications and benchmarks (prospecting, feature characterisation, etc.) 

\subsubsection{Expert-informed Few-shot learning}

\begin{figure}[h!]
    \centering
    \includegraphics[width=\textwidth]{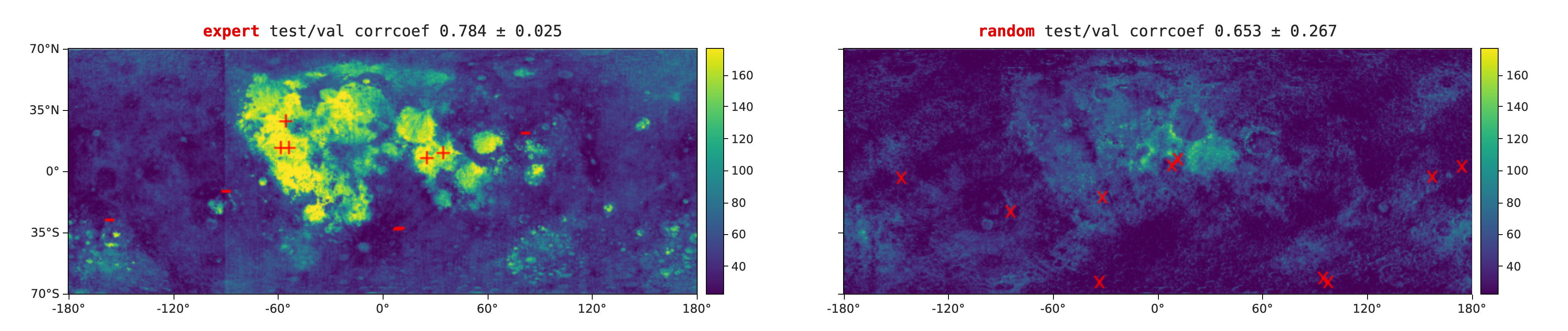}
    \caption{Random forest models trained with expert location selection vs. random selection. The red plus and minus signs on the left-hand chart indicate the expert's positive and negative choices in one of the expert runs. Red crosses on the right map show the random choices on one of the random runs. You can compare this with the \chem{TiO_2} ground truth map in Fig. \ref{fig:downstream_tasks}.}
    \label{fig:tio2_expert}
\end{figure}

Furthermore, we evaluate LunarFM in a more realistic scenario, in which we have access to only a handful of expert-curated data points to train a downstream regression model. To this end, we use the recent work by \cite{diaz2025descriptive}, in which the authors developed a descriptive/conceptual model to map areas on the Moon with high concentrations of ilmenite (\chem{TiO_2}). These high-titanium areas intersect with approximately 800 half-degree lunar chips. From these, we randomly select 5 chips as positive examples and then we also select 5 chips with very low percentile of \chem{TiO_2} concentration as negative examples. In this way, we approximate the scenario in which an expert selects 10 chips with known (very high and very low) mineral concentrations. We then train a random forest that takes LunarFM chip embeddings as inputs and predicts \chem{TiO_2} concentration across the lunar surface. We repeat the experiment 20 times and report the mean and standard deviation of the correlation coefficient between the ground truth and the predictions. In parallel, we perform a similar experiment, but instead of using an expert-curated (and thus highly informative) training dataset, we randomly select 10 chips from the Moon's surface. \Cref{fig:tio2_expert} shows the results. However, when combined with expert advice on selecting high-quality training locations, the downstream embedding training yields substantially better predictions, with 10× lower variability.

\subsubsection{Geological-Unit Classification}
\label{sec:geological-unit-classification}

As an additional downstream evaluation, we test whether LunarFM embeddings contain information relevant to discrete geological mapping. We consider the task of predicting the dominant geologic units in the USGS Unified Geologic Map of the Moon 1:5M \citep{fortezzo2020release} associated with each chip. The USGS Unified Geologic Map uses 49 classes to encode the inferred geologic age, the geologic origin, and the geomorphology of the surface. This is a fine-grained multi-class classification problem with substantial class imbalance, strong spatial structure, and moderately subjective ground truth that is based on expert consensus. Importantly, the classifier is trained only on the 768-dimensional LunarFM chip embeddings; no imagery, coordinates, or geological labels are used as additional inputs for this downstream task.

We train a histogram-based gradient boosting classifier on the chip-level embeddings and evaluate three distinct spatial train/validation/test strategies to test various degrees of generalisation. The first is the default diagonal band split used elsewhere in the paper, which is designed to reduce local spatial leakage while maintaining broad geographic coverage in each split. The second holds out the easternmost 20\% of longitudes as a contiguous test region. The third holds out the northernmost 20\% of latitudes, corresponding to a polar-cap extrapolation setting. These latter two splits are deliberately much harder tests of spatial generalisation because the model must transfer from one contiguous geographic region to another.

Performance is evaluated using top-1 accuracy, macro-averaged F1 score (macro-F1), and top-3 accuracy. Top-1 and top-3 accuracy measure whether the correct geological unit is the highest-ranked or among the three highest-ranked predictions, respectively, while macro-F1 computes the F1 score independently for each class and averages across classes, giving equal weight to rare and frequent geological units.

\begin{figure}[h!]
    \centering
    \includegraphics[width=\textwidth]{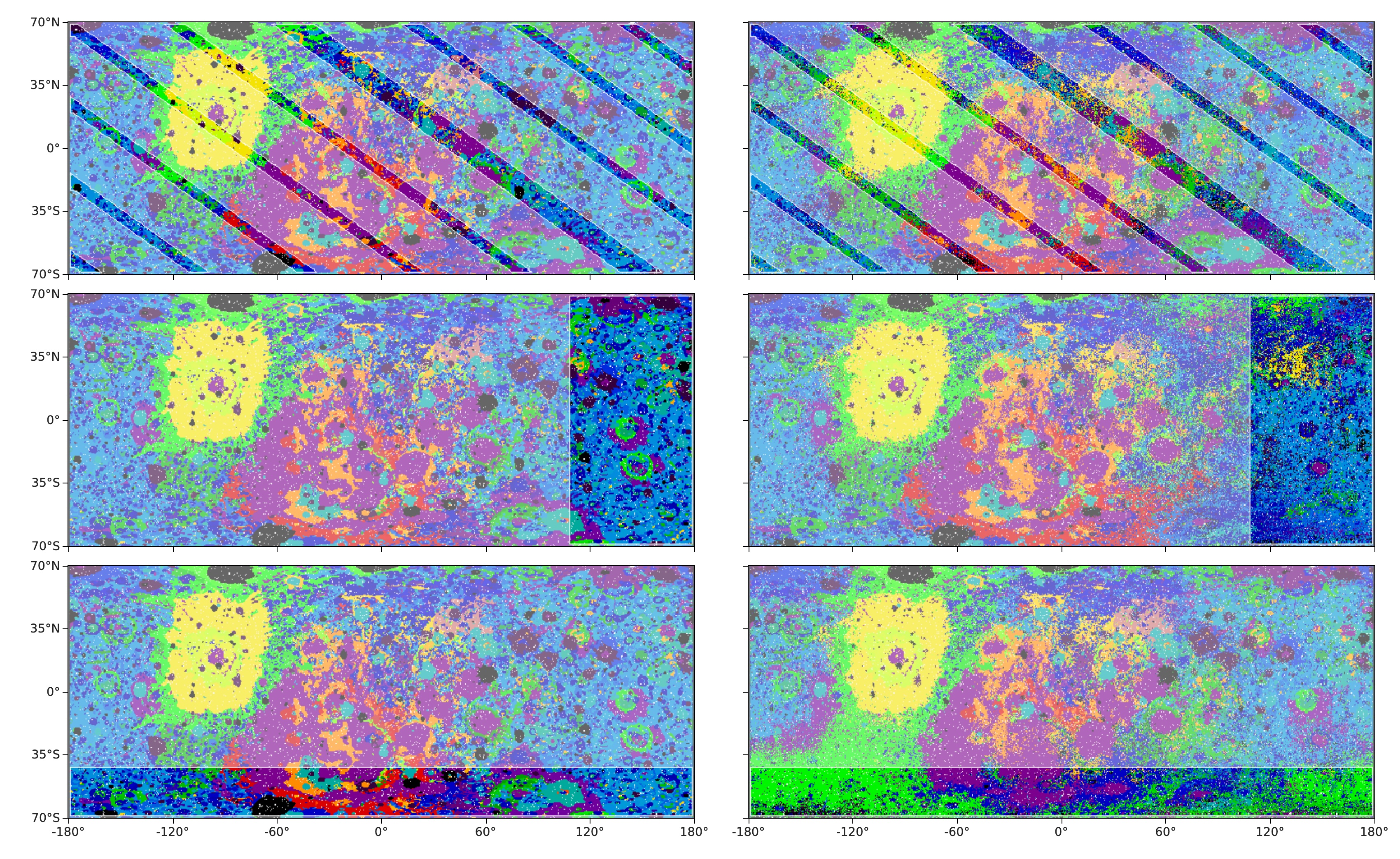}
    \caption{USGS 49-band geological-unit classification maps. Left: ground truth. Centre: full-surface predictions from the gradient-boosted classifier trained on the LunarFM embeddings. Right: predictions shown only over the corresponding test region.}
    \label{fig:geology_prediction_maps}
\end{figure}

\begin{figure}[h!]
        \centering
    \includegraphics[width=0.85\textwidth]{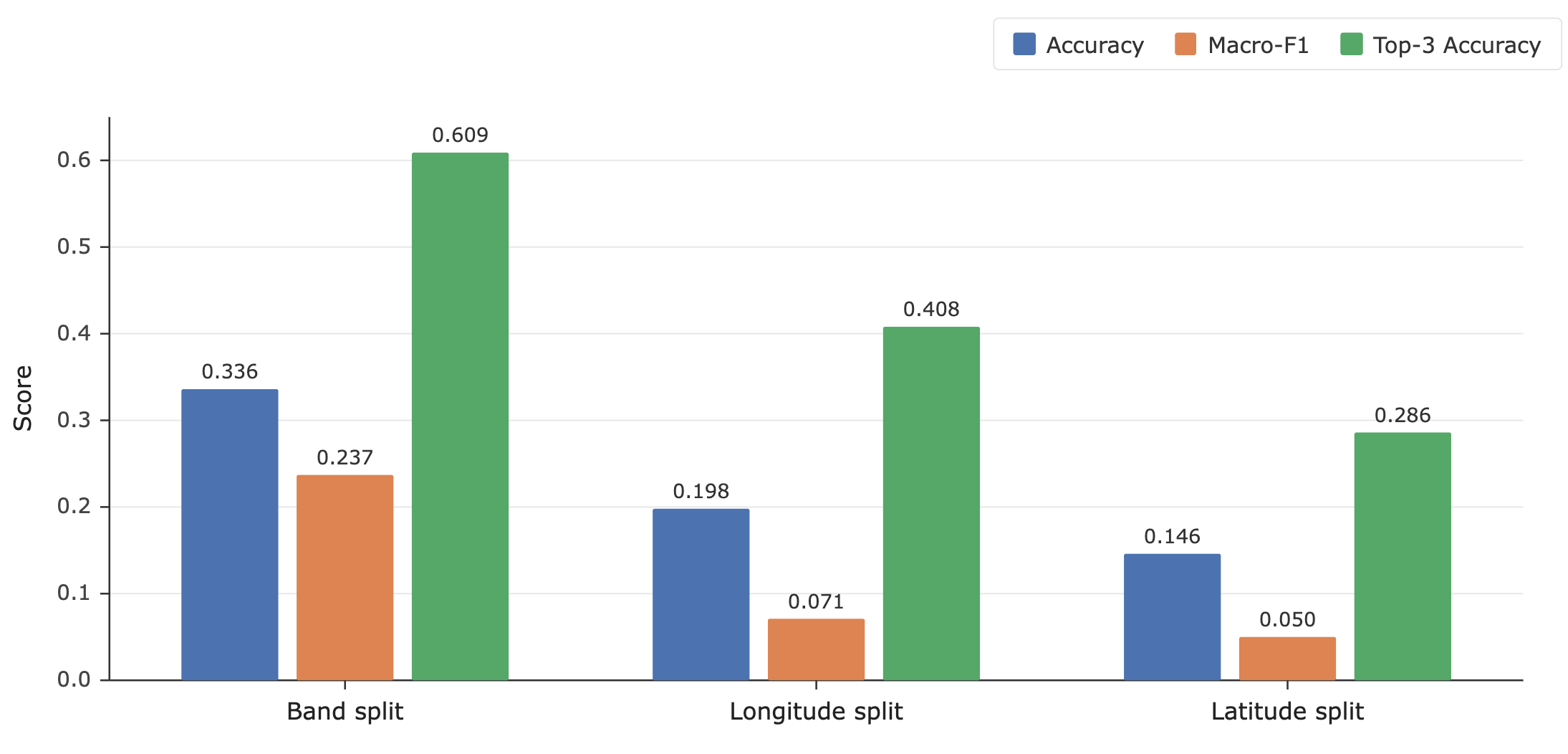}
    \caption{USGS 49-band geological-unit classification performance using frozen LunarFM embeddings and a gradient-boosted classifier under three spatial splitting strategies. The default band split gives the strongest performance, while contiguous longitude and latitude holdouts expose substantially harder spatial generalisation regimes, which is not surprising given the Moon's spatially heterogeneous surface.}
    \label{fig:geology_classification_metrics}
\end{figure}

The results in \Cref{fig:geology_prediction_maps} show that LunarFM embeddings are predictive of geological units under the standard band split, for which the classifier reaches 33.6\% top-1 accuracy, 23.7\% macro-F1, and 60.9\% top-3 accuracy (\Cref{fig:geology_classification_metrics}). The relatively high top-3 accuracy is notable given the fine-grained 49-class target space and the ambiguity expected between neighbouring or related geological units. However, when the test set is defined as a contiguous longitude strip, performance drops to 19.8\% accuracy, 7.1\% macro-F1, and 40.8\% top-3 accuracy. The latitude split proves harder still, with 14.6\% accuracy, 5.0\% macro-F1, and 28.6\% top-3 accuracy.

These results have two implications. First, they provide evidence that LunarFM embeddings encode geologically meaningful information beyond simple reconstruction quality, i.e., a lightweight classifier trained on frozen embeddings can recover non-trivial geological-unit structure. 
Second, they show that evaluation protocol matters. The default band split is useful as a spatially disjoint benchmark, but it still provides broad latitudinal and longitudinal coverage during training. 
In contrast, the contiguous longitude and latitude holdouts introduce stronger distribution shift, particularly for geological units concentrated in specific regions. The sharp drop in macro-F1 under these splits indicates that some classes are poorly represented, or even absent, in the corresponding training regions. Thus, the band-split result should be interpreted as evidence of useful geological information in the embeddings, rather than as an estimate of performance under arbitrary geographic extrapolation.

Overall, the experiment supports the use of LunarFM embeddings as a compact representation for downstream geological analysis, while also highlighting the need for task-specific evaluation protocols when the intended application involves extrapolation to geologically and environmentally distinct regions, such as, for example, the lunar poles.

\section{Discussion, Limitations \& Conclusion} \label{sec:discussion}
The evaluations presented above provide encouraging evidence that multimodal self-supervised pretraining produces representations with meaningful structure for lunar science applications.
The reconstruction results indicate that the model learns both intra-modality and cross-modality correlations, and even when a full modality is withheld at inference, the model produces plausible reconstructions by drawing on complementary information from the remaining inputs. However, reconstruction quality varies substantially across modalities and examples. The Mini-RF error likely reflects the presence of NaN pixels in the training data and noise in radar backscatter. We also note that Diviner Rock Abundance (RA) occasionally exhibits unexpected reconstruction behaviour, with outputs that appear to track Diviner temperature products rather than rock abundance, likely due to the strong correlation between these two inputs.
The embedding analyses provide qualitative evidence that the representation space organises along physically meaningful dimensions. PCA components exhibit spatially coherent patterns broadly consistent with known geological terrain types. UMAP projections reveal a continuous manifold with differentiated subregions, and colouring by \chem{TiO_2} concentration identifies a distinct cluster of high-titanium chips within this space. Unsupervised k-means clustering yields spatially contiguous groupings without geographic constraints, suggesting that the embeddings encode large-scale surface variability in a geographically coherent manner. We note that evaluating learned representations in a domain with scarce and spatially irregular ground truth is inherently challenging. Identifying resource-relevant structure in an unsupervised setting proved non-trivial: while high-\chem{TiO_2} regions form a discernible subregion in embedding space, separating specific resource signatures from broader terrain patterns requires targeted evaluation protocols that go beyond what is established here.
\paragraph{Downstream Use Cases.}
The downstream experiments are presented as illustrative use cases that demonstrate the practical potential of the LunarFM embeddings, rather than as exhaustive benchmarks. 
The mineral mapping experiment (Section~\ref{sec:downstream}) yields spatially coherent predictions of mineral concentrations including \chem{FeO}, \chem{TiO_2}, \chem{MgO}, and \chem{Al_2O_3}, with reasonable agreement between predicted and reference maps, illustrating that the embeddings are directly useful for regression-based compositional mapping even with limited supervision.
In a more operationally realistic scenario, we simulated the workflow of an expert selecting a small number of reference locations with known high and low \chem{TiO_2} concentrations. Using only 10 expert-selected chips for training, the random forest achieves a correlation of $0.784 \pm 0.025$ with the ground-truth map, compared to $0.653 \pm 0.267$ when the same number of chips is selected at random. The substantially lower variance in the expert condition highlights an important practical finding: the value of the embeddings is amplified when combined with domain knowledge in sample selection, reflecting a realistic workflow for targeted resource assessment where a geologist contributes a small number of high-confidence annotations and the embedding space generalises across the remainder. 
Querying the embedding space using the L2 distance retrieves chips that are visually and geophysically similar across multiple modalities. This capability could support prospecting workflows in which an analyst identifies a region of interest and seeks analogous surface conditions elsewhere on the Moon, without requiring explicit spectral or compositional criteria.
\paragraph{Spatial Generalisation.}
The geological-unit classification experiment further highlights the importance of spatial evaluation protocols. LunarFM embeddings support non-trivial prediction of fine-grained geological units under the default band split, suggesting that the representation captures physically and geologically meaningful structure. However, performance drops substantially when the test region is defined as a contiguous longitude or latitude holdout. This indicates that some downstream tasks may be sensitive to geographic distribution shift, especially when target classes are spatially localised or absent from the training region. Future benchmarks should therefore report not only random or interleaved spatial splits, but also harder contiguous-region holdouts that better approximate extrapolation to geologically distinct areas.
\paragraph{Limitations.}
LunarFM is a proof-of-principle framework, and several aspects reflect both the practical constraints under which it was developed and the fundamental challenges of the lunar domain. 
The dataset is restricted to $\pm$70° latitude due to gaps in modality coverage at high latitudes and distinct illumination-related challenges, thereby excluding the lunar south pole, i.e., the primary region of interest for water-ice prospecting operations. 
The Mini-RF data modality contains substantial NaN regions from incomplete orbital coverage; these NaN regions are masked during training, which limits the model's ability to fully exploit radar backscatter information and, in turn, contributes to the substantially higher reconstruction error observed for this modality. Chip-level embeddings collapse a 0.5°$\times$0.5° surface area into a single vector, corresponding to roughly 15~km in the north-south direction, with the east-west extent varying from $\sim$15~km at the equator to $\sim$5~km near 70° latitude. This varying footprint size implies that embeddings represent different physical extents at different latitudes, complicating direct comparisons across the coverage area. Given the non-inclusion of high-resolution products, such as LROC Narrow Angle Camera (NAC) imagery, the current resolution may be insufficient for applications requiring fine or spatially consistent discrimination, such as landing-site characterisation or sub-km surface feature analysis. Leveraging the patch-level embeddings from the same encoder is a natural approach to addressing higher-resolution downstream use cases; however, variability in spatial resolution across data modalities and latitudes poses additional challenges. 
More broadly, rigorous evaluation of representation quality in the absence of large labelled datasets is an open methodological challenge in planetary science. The PCA and UMAP analyses provide interpretable visualisations but do not constitute formal proof of geophysical fidelity, and establishing quantitative evaluation protocols is an important direction for follow-up work. 
Finally, the model architecture and hyperparameters were set based on prior work and practical constraints rather than domain-specific optimisation, and both have significant room for refinement in future iterations.
\paragraph{Conclusion \& Outlook.}
LunarFM provides a reproducible multimodal latent representation for the lunar ML community, integrating six physically distinct remote-sensing modalities into a shared representation space. The presented use cases demonstrate that these embeddings can be useful for low-label mineral mapping and similarity-based retrieval, and the open release of the dataset, model, embeddings, and notebooks is intended to lower the barrier for the broader community to explore and extend these capabilities, refine the framework, and identify the most promising directions for future work. Future work can build on this foundation in several directions. Extending geographic coverage to include polar regions would extend the framework to scientifically and operationally important areas of the Moon. Incorporating NAC-scale imagery, as well as other input datasets such as terrain elevation derivatives (slope, aspect, roughness) or hyperspectral imagery, and moving toward temporally-indexed inputs might better capture dynamics (e.g., thermal) and mission-dependent observation conditions. 
On the modelling side, alternative self-supervised objectives, such as contrastive learning or joint-embedding predictive architectures, may better disentangle correlated modalities such as the Diviner thermophysical channels, and could be explored in future iterations of the framework. 
Developing principled benchmarks for representation quality in low-label planetary science settings is a priority, including reference against unimodal and raw-feature approaches to quantify what multimodal pretraining specifically contributes. If the embeddings prove reliable for specific resource-oriented tasks, a natural next step is integrating them into broader analytical pipelines, coupling similarity search with expert-in-the-loop workflows for rapid site screening, or connecting the representation space to language-based retrieval for human-interpretable query interfaces. We view LunarFM not as a finished product but as an infrastructure layer, one that we hope future missions, datasets, and scientific questions can continue to build upon.

\section{Acknowledgements}

This work has been enabled by Frontier Development Lab Europe (https://fdleurope.org), a public/private partnership between the European Space Agency (ESA), the Luxembourg Space Agency, Datarock, Trillium Technologies,
the University of Oxford and leaders in commercial AI supported by Google Cloud, SCAN and NVIDIA, developing open science for all Humankind. Furthermore, the authors thank David Briguglio, Katherine Hadler, Mike Heyns, Alex Maritati, Thomas Schaap, and James Parr for their guidance, discussions, and feedback throughout this research.

\newpage

\bibliography{iclr2026_conference}

@article{fortezzo2020release,
  title={Release of the digital unified global geologic map of the moon at 1: 5,000,000},
  author={Fortezzo, Corey M. and Spudis, Paul D. and Harrel, Shannon L.},
  journal={51st Lunar and Planetary Science Conference},
  year={2020}
}

@article{prasad2026moonstone,
  title={Moonstone: A Multimodal Foundation Model and Benchmark for Lunar Remote Sensing},
  author={Prasad, Ayush and Mazumder, Swarnalee},
  journal={arXiv preprint arXiv:2607.03644},
  year={2026}
}

@article{fang2026domain,
  title={A Domain-Specific Vision Foundation Model for Mars: Self-Supervised Learning for Planetary-Scale Science Discovery},
  author={Fang, Jichao and Luo, Wei and Huang, Qunying and Zhang, Lei and Phillips, Michael and Seethi, Venkata Devesh Reddy and Giannakis, Iraklis},
  journal={Authorea Preprints},
  year={2026},
  publisher={Authorea}
}

@inproceedings{inal2026llava,
  title={LLaVA-LE: Large Language-and-Vision Assistant for Lunar Exploration},
  author={Inal, Gokce and Navard, Pouyan and Yilmaz, Alper},
  booktitle={Proceedings of the IEEE/CVF Conference on Computer Vision and Pattern Recognition},
  pages={10251--10261},
  year={2026}
}

@inproceedings{purohit2026momo,
  title={MOMO: Mars Orbital Model Foundation Model for Mars Orbital Applications},
  author={Purohit, Mirali and Gajera, Bimal and Mehta, Irish and Tokas, Bhanu and Adler, Jacob and Lu, Steven and Dickenshied, Scott and Diniega, Serina and Bue, Brian and Rebbapragada, Umaa and others},
  booktitle={Proceedings of the IEEE/CVF Conference on Computer Vision and Pattern Recognition},
  pages={27772--27782},
  year={2026}
}

@article{prettyman2006_elemental_composition_lunar_prospector,
  title        = {{Elemental composition of the lunar surface: {A}nalysis of gamma ray spectroscopy data from {L}unar Prospector}},
  author       = {Prettyman, T. H. and Hagerty, J. J. and Elphic, R. C. and Feldman, W. C. and Lawrence, D. J. and McKinney, G. W. and Vaniman, D. T.},
  journal      = {Journal of Geophysical Research: Planets},
  volume       = {111},
  number       = {E12},
  year         = {2006},
  doi          = {10.1029/2005JE002656},
  url          = {https://agupubs.onlinelibrary.wiley.com/doi/full/10.1029/2005JE002656}
}

@article{jakubik2025terramind,
  title={{{T}erra{M}ind: {L}arge-Scale Generative Multimodality for {E}arth Observation}},
  author={Jakubik, Johannes and Yang, Felix and Blumenstiel, Benedikt and Scheurer, Erik and Sedona, Rocco and Maurogiovanni, Stefano and Bosmans, Jente and Dionelis, Nikolaos and Marsocci, Valerio and Kopp, Niklas and Ramachandran, Rahul and Fraccaro, Paolo and Brunschwiler, Thomas and Gabriele, Cavallaro and Bernabe-Moreno, Juan and Longépé, Nicolas},
  journal={IEEE/CVF International Conference on Computer Vision (ICCV)},
  year={2025}
}

@inproceedings{tsenggalileo,
  title={{Galileo: Learning Global \& Local Features of Many Remote Sensing Modalities}},
  author={Tseng, Gabriel and Fuller, Anthony and Reil, Marlena and Herzog, Henry and Beukema, Patrick and Bastani, Favyen and Green, James R and Shelhamer, Evan and Kerner, Hannah and Rolnick, David},
  booktitle={Forty-second International Conference on Machine Learning},
  year={2025}
}

@article{neish2011nature,
  title={{The nature of lunar volatiles as revealed by Mini-RF observations of the LCROSS impact site}},
  author={Neish, CD and Bussey, DBJ and Spudis, P and Marshall, W and Thomson, BJ and Patterson, GW and Carter, LM},
  journal={Journal of Geophysical Research: Planets},
  volume={116},
  number={E1},
  year={2011},
  publisher={Wiley Online Library}
}

@inproceedings{astruc2025anysat,
  title={{AnySat: One Earth Observation Model for Many Resolutions, Scales, and Modalities}},
  author={Astruc, Guillaume and Gonthier, Nicolas and Mallet, Clement and Landrieu, Loic},
  booktitle={Proceedings of the Computer Vision and Pattern Recognition Conference},
  pages={19530--19540},
  year={2025}
}

@inproceedings{bachmann2022multimae,
  title={{Multimae: Multi-modal multi-task masked autoencoders}},
  author={Bachmann, Roman and Mizrahi, David and Atanov, Andrei and Zamir, Amir},
  booktitle={European Conference on Computer Vision},
  pages={348--367},
  year={2022},
  organization={Springer}
}

@article{diaz2025descriptive,
  title={{Descriptive models for lunar High-Ti deposits}},
  author={Diaz, Abigail Calzada and Keszthelyi, Laszlo},
  journal={Acta Astronautica},
  volume={226},
  pages={375--384},
  year={2025},
  publisher={Elsevier}
}

@article{pieters2009,
  title={{Character and spatial distribution of OH/H2O on the surface of the Moon seen by M3 on Chandrayaan-1}},
  author={Pieters, Carle M and Goswami, JN and Clark, RN and Annadurai, M and Boardman, J and Buratti, B and Combe, J-P and Dyar, MD and Green, R and Head, JW and others},
  journal={science},
  volume={326},
  number={5952},
  pages={568--572},
  year={2009},
  publisher={American Association for the Advancement of Science}
}

@article{Colaprete2010,
  title={{Detection of water in the LCROSS ejecta plume}},
  author={Colaprete, Anthony and Schultz, Peter and Heldmann, Jennifer and Wooden, Diane and Shirley, Mark and Ennico, Kimberly and Hermalyn, Brendan and Marshall, William and Ricco, Antonio and Elphic, Richard C and Goldstein, David and Summy, Dustin and Bart, Gwendolyn D. and Asphaug, Erik and Korycansky, Don and Landis, David and Sollitt, Luke},
  journal={science},
  volume={330},
  number={6003},
  pages={463--468},
  year={2010},
  publisher={American Association for the Advancement of Science}
}

@article{wang_self-supervised_2022,
	title        = {{Self-{Supervised} {Learning} in {Remote} {Sensing}: {A} review}},
	author       = {Wang, Yi and Albrecht, Conrad M. and Braham, Nassim Ait Ali and Mou, Lichao and Zhu, Xiao Xiang},
	year         = 2022,
	journal      = {IEEE Geoscience and Remote Sensing Magazine},
	volume       = 10,
	number       = 4,
	pages        = {213--247},
	doi          = {10.1109/MGRS.2022.3198244},
	issn         = {2168-6831},
	note         = {Conference Name: IEEE Geoscience and Remote Sensing Magazine}
}

@article{tao_self-supervised_2023,
	title        = {{Self-supervised remote sensing feature learning: {Learning} {Paradigms}, {Challenges}, and {Future} {Works}}},
	author       = {Tao, Chao and Qi, Ji and Guo, Mingning and Zhu, Qing and Li, Haifeng},
	year         = 2023,
	journal      = {IEEE Transactions on Geoscience and Remote Sensing},
	volume       = 61,
	pages        = {1--26},
	doi          = {10.1109/TGRS.2023.3276853},
	issn         = {0196-2892, 1558-0644},
	note         = {arXiv:2211.08129 [cs]}
}

@misc{fuller2023cromaremotesensingrepresentations,
      title={{CROMA: Remote Sensing Representations with Contrastive Radar-Optical Masked Autoencoders}}, 
      author={Anthony Fuller and Koreen Millard and James R. Green},
      year={2023},
      eprint={2311.00566},
      archivePrefix={arXiv},
      primaryClass={cs.CV},
      url={https://arxiv.org/abs/2311.00566}, 
}

@misc{xiong2025neuralplasticityinspiredmultimodalfoundation,
      title={{Neural Plasticity-Inspired Multimodal Foundation Model for Earth Observation}}, 
      author={Zhitong Xiong and Yi Wang and Fahong Zhang and Adam J. Stewart and Joëlle Hanna and Damian Borth and Ioannis Papoutsis and Bertrand Le Saux and Gustau Camps-Valls and Xiao Xiang Zhu},
      year={2025},
      eprint={2403.15356},
      archivePrefix={arXiv},
      primaryClass={cs.CV},
      url={https://arxiv.org/abs/2403.15356}, 
}

@article{sato2017tio2,
  title={{Lunar mare {TiO2} abundances estimated from UV/Vis reflectance}},
  author={Sato, Hiroyuki and Robinson, Mark S and Lawrence, Samuel J and Denevi, Brett W and Hapke, Bruce and Jolliff, Bradley L and Hiesinger, Harald},
  journal={Icarus},
  volume={296},
  pages={216--238},
  year={2017},
  publisher={Elsevier}
}

@misc{sander2025moonsfacessingleunified,
      title={{The Moon's Many Faces: A Single Unified Transformer for Multimodal Lunar Reconstruction}}, 
      author={Tom Sander and Moritz Tenthoff and Kay Wohlfarth and Christian Wöhler},
      year={2025},
      eprint={2505.05644},
      archivePrefix={arXiv},
      primaryClass={cs.CV},
      url={https://arxiv.org/abs/2505.05644}, 
}

@article{cuervo2025moon,
  title={{Moon photovoltaics utilizing lunar regolith and halide perovskites}},
  author={Cuervo-Ortiz, Juli{\'a}n Mauricio and Palomares, Juan Carlos Gin{\'e}s and Ozen, Sercan and H{\"a}rtel, Marlene and Sarisozen, Sema and Dittwald, Alina and Kourkafas, Georgios and Castro-M{\'e}ndez, Andr{\'e}s-Felipe and Pe{\~n}a-Camargo, Francisco and Seid, Biruk Alebachew and others},
  journal={Device},
  volume={3},
  number={7},
  year={2025},
  publisher={Elsevier}
}

@article{sargeant2020hydrogen,
  title={{Hydrogen reduction of ilmenite: Towards an in situ resource utilization demonstration on the surface of the Moon}},
  author={Sargeant, HM and Abernethy, FAJ and Barber, SJ and Wright, IP and Anand, M and Sheridan, S and Morse, A},
  journal={Planetary and Space Science},
  volume={180},
  pages={104751},
  year={2020},
  publisher={Elsevier}
}

@INPROCEEDINGS{hedrick2023reemoon,
  author={Hedrick, Gabrielle},
  booktitle={2023 IEEE Aerospace Conference}, 
  title={{Towards Mining Rare Earth Elements on the Moon}}, 
  year={2023},
  volume={},
  number={},
  pages={1-11},
  doi={10.1109/AERO55745.2023.10116027}}

@article{https://doi.org/10.1029/2018JE005592,
author = {Robbins, Stuart J.},
title = {{A New Global Database of Lunar Impact Craters >1–2 km: 1. Crater Locations and Sizes, Comparisons With Published Databases, and Global Analysis}},
journal = {Journal of Geophysical Research: Planets},
volume = {124},
number = {4},
pages = {871-892},
doi = {https://doi.org/10.1029/2018JE005592},
url = {https://agupubs.onlinelibrary.wiley.com/doi/abs/10.1029/2018JE005592},
eprint = {https://agupubs.onlinelibrary.wiley.com/doi/pdf/10.1029/2018JE005592},
year = {2019}
}

@article{speyerer2023precise,
  title={{Precise mapping of the Moon with the Clementine Ultraviolet/Visible Camera}},
  author={Speyerer, Emerson J and Robinson, Mark S and Boyd, Aaron and Silva, Victor H and Lawrence, Samuel},
  journal={Icarus},
  volume={398},
  pages={115506},
  year={2023},
  publisher={Elsevier}
}

@article{powell2023diviner,
author = {Powell, T. M. and Horvath, T. and Robles, V. Lopez and Williams, J.-P. and Hayne, P. O. and Gallinger, C. L. and Greenhagen, B. T. and McDougall, D. S. and Paige, D. A.},
title = {{High-Resolution Nighttime Temperature and Rock Abundance Mapping of the Moon Using the Diviner Lunar Radiometer Experiment With a Model for Topographic Removal}},
journal = {Journal of Geophysical Research: Planets},
volume = {128},
number = {2},
pages = {e2022JE007532},
doi = {https://doi.org/10.1029/2022JE007532},
url = {https://agupubs.onlinelibrary.wiley.com/doi/abs/10.1029/2022JE007532},
eprint = {https://agupubs.onlinelibrary.wiley.com/doi/pdf/10.1029/2022JE007532},
note = {e2022JE007532 2022JE007532},
year = {2023}
}

@article{goossens2020grail,
author = {Goossens, S. and Sabaka, T. J. and Wieczorek, M. A. and Neumann, G. A. and Mazarico, E. and Lemoine, F. G. and Nicholas, J. B. and Smith, D. E. and Zuber, M. T.},
title ={ {High-Resolution Gravity Field Models from GRAIL Data and Implications for Models of the Density Structure of the Moon's Crust}},
journal = {Journal of Geophysical Research: Planets},
volume = {125},
number = {2},
pages = {e2019JE006086},
doi = {https://doi.org/10.1029/2019JE006086},
url = {https://agupubs.onlinelibrary.wiley.com/doi/abs/10.1029/2019JE006086},
eprint = {https://agupubs.onlinelibrary.wiley.com/doi/pdf/10.1029/2019JE006086},
note = {e2019JE006086 10.1029/2019JE006086},
year = {2020}
}

@article{nozette2010lunar,
  title={The lunar reconnaissance orbiter miniature radio frequency (Mini-RF) technology demonstration},
  author={Nozette, Stewart and Spudis, Paul and Bussey, Ben and Jensen, Robert and Raney, Keith and Winters, Helene and Lichtenberg, Christopher L and Marinelli, William and Crusan, Jason and Gates, Michele and others},
  journal={Space Science Reviews},
  volume={150},
  number={1},
  pages={285--302},
  year={2010},
  publisher={Springer}
}
\bibliographystyle{iclr2026_conference}

\newpage
\appendix

\section{Dataset sources and splits}
\label{app:data_sources}

This appendix details the source products used to construct the LunarChips dataset and the additional datasets used in downstream experiments. All products were obtained from NASA's Planetary Data System (PDS) or from mission-associated lunar map repositories, and were subsequently reprojected, co-registered, resampled, and tiled into the common 0.5$^\circ \times$ 0.5$^\circ$ chip grid described in Section~\ref{sec:technical-setup}. The representation-learning stage uses the six primary sensing modalities described in Table~\ref{tab:lunar_instruments}; downstream labels and auxiliary layers are used only for evaluation and illustrative applications.

\subsection{Primary input modalities used for LunarFM pretraining} \label{app-sec:input-data}

\paragraph{LRO LROC WAC Hapke Photometric Maps (7 channels).}
The multispectral optical input is derived from the Lunar Reconnaissance Orbiter Camera (LROC) Wide Angle Camera (WAC) Hapke-normalised mosaic product, available through the LROC data portal. We use the seven-band UV--visible mosaic spanning 321--689~nm. This product provides photometrically normalised reflectance and is based on the Hapke parameter mapping framework described by \cite{sato2017tio2}. In LunarFM, these bands form the main multispectral optical modality.

\paragraph{LRO LOLA topography (1 channel).}
The topographic input is derived from the Lunar Orbiter Laser Altimeter (LOLA) gridded digital elevation model (DEM) distributed through the LRO/LOLA archive. LOLA provides the geodetic and topographic reference framework for the Moon. We use the global gridded elevation product as the single-channel topography modality.

\paragraph{LRO Diviner thermophysical products (3 channels).}
The thermal modality includes Diviner Lunar Radiometer-derived products. In particular, we use global maps of regolith temperature, bolometric temperature, and rock abundance derived from long-term nighttime anisothermality observations. These products follow the updated global thermophysical mapping framework of \cite{powell2023diviner}. In the model implementation, Diviner-derived inputs are grouped into thermophysical channels, with rock abundance treated separately in preprocessing/model grouping.

\paragraph{LRO Mini-RF radar products (2 channels).}
The radar modality is derived from LRO Mini-RF products distributed through the PDS Geosciences Node. Mini-RF is a synthetic aperture radar instrument designed to characterise lunar surface roughness, blockiness, and radar backscattering properties \citep{nozette2010lunar}. The LunarFM input stack uses two radar-derived channels, namely the circular polarisation ratio (CPR) and the first Stokes parameter (S1), both resampled to the common chip grid.

\paragraph{GRAIL gravity-derived products (4 channels).}
The gravity modality comprises products from the Gravity Recovery and Interior Laboratory (GRAIL) mission. These layers were assembled from GRAIL-derived gravity products, including free-air gravity anomaly, Bouguer anomaly, and additional gravity-related fields used as proxies for subsurface structure. The underlying gravity-field models are described by \cite{goossens2020grail}. In our data stack, four gravity-related channels are included: free-air anomaly, Bouguer anomaly, disturbance, and an uncertainty layer.

\paragraph{Clementine UVVIS reflectance/albedo (1 channel).}
Clementine UVVIS products provide a multispectral global view of surface reflectance. For LunarFM, we include the 750 nm wavelength Clementine-derived reflectance channel, resampled to the common grid. The instrument and mapping quality are discussed in more recent reprocessing and geodetic reassessment work by \cite{speyerer2023precise}.

\subsection{Additional datasets used for downstream tasks and evaluation} \label{app-sec:downstream-data}

\paragraph{Lunar Prospector gamma-ray spectroscopy compositional maps.}
For mineral and elemental downstream tasks, LunarChips includes global compositional maps derived from Lunar Prospector Gamma-Ray Spectrometer (GRS) measurements. These maps provide estimates of major-element abundances and are based on the analysis framework of \cite{prettyman2006_elemental_composition_lunar_prospector}. In our experiments, these layers are used only as supervised learning targets for downstream regression tasks.

\paragraph{WAC-derived ilmenite map.}
For downstream experiments related to ilmenite, we use the LROC WAC TiO$_2$ abundance product derived from UV/Vis reflectance. This map follows the methodology of \cite{sato2017tio2} and, in this work, is used as a label dataset for supervised learning TiO$_2$ mapping experiments.

\paragraph{Crater annotations.}
For crater-focused analyses, LunarChips incorporates crater annotations from the global lunar crater database of \cite{https://doi.org/10.1029/2018JE005592} (craters $\geq$ 1–2 km in diameter). This database provides global crater locations and sizes and is also used for downstream supervised learning tasks.

\paragraph{Expert-curated high-TiO$_2$ region delineations.}
For the few-shot resource-screening experiment, we additionally use the high-TiO$_2$ region delineations proposed by \citet{diaz2025descriptive}. In this work, these delineations are treated as an expert-curated reference set for simulating realistic low-label expert-guided sampling. These data are used exclusively in downstream experiments.

\paragraph{USGS geological-unit labels.}
For the geological-unit classification experiment, we use chip-aligned labels derived from USGS lunar geological mapping products \citep{fortezzo2020release}. Each chip is assigned the dominant detailed geological unit within its footprint, producing a multi-class classification target for evaluating whether frozen LunarFM embeddings encode mappable geological structure. These labels are used only for downstream evaluation and are not used during self-supervised pretraining.

\subsection{Dataset Splits} \label{app-sec:data-split}
\begin{figure}[h!]
    \centering
    \begin{subfigure}[t]{0.64\textwidth}
        \centering
        \includegraphics[width=\textwidth]{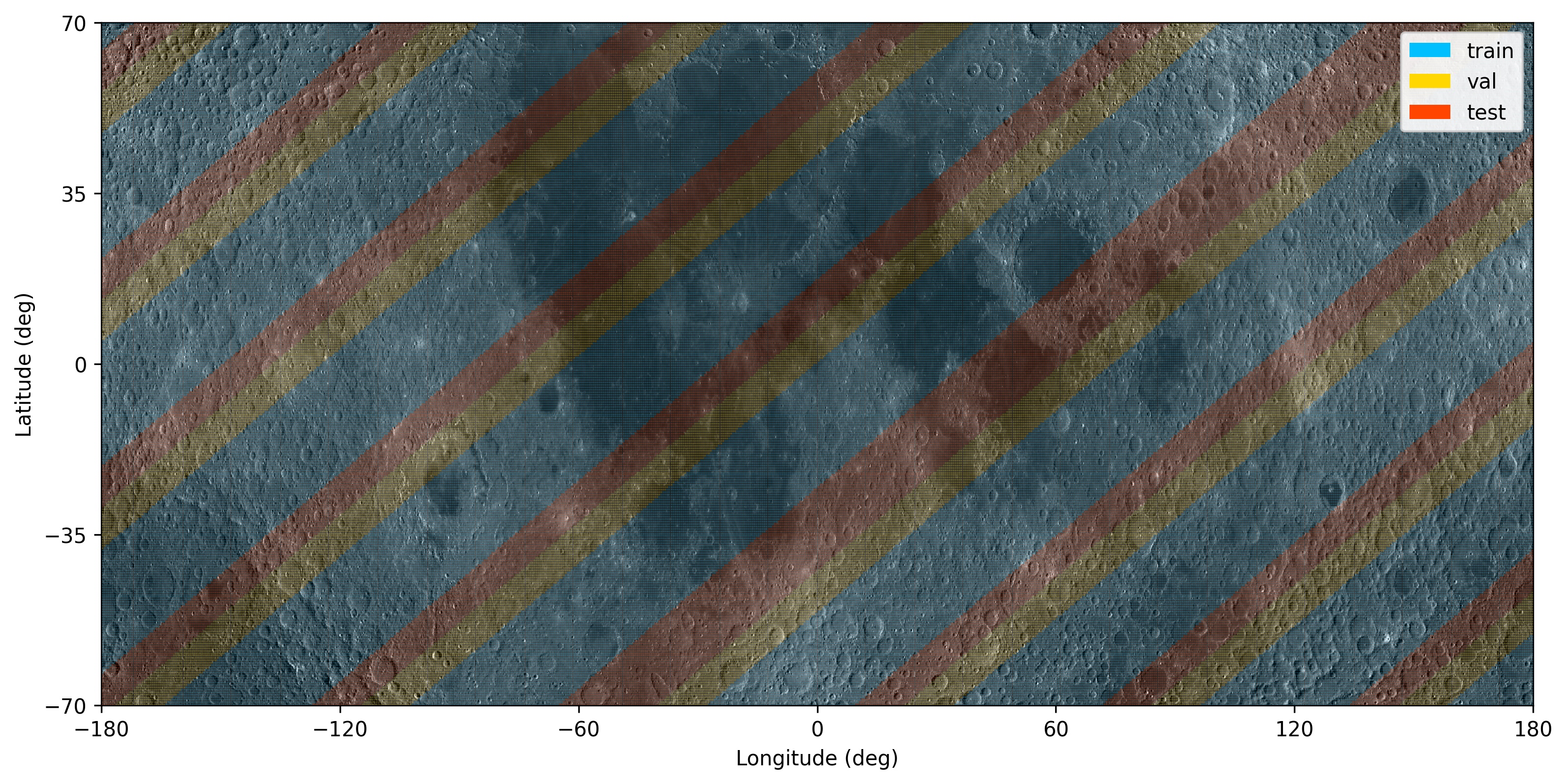}
        \caption{Global view.}
        \label{fig:bandsplit_global}
    \end{subfigure}\hfill
    \begin{subfigure}[t]{0.34\textwidth}
        \centering
        \includegraphics[width=\textwidth]{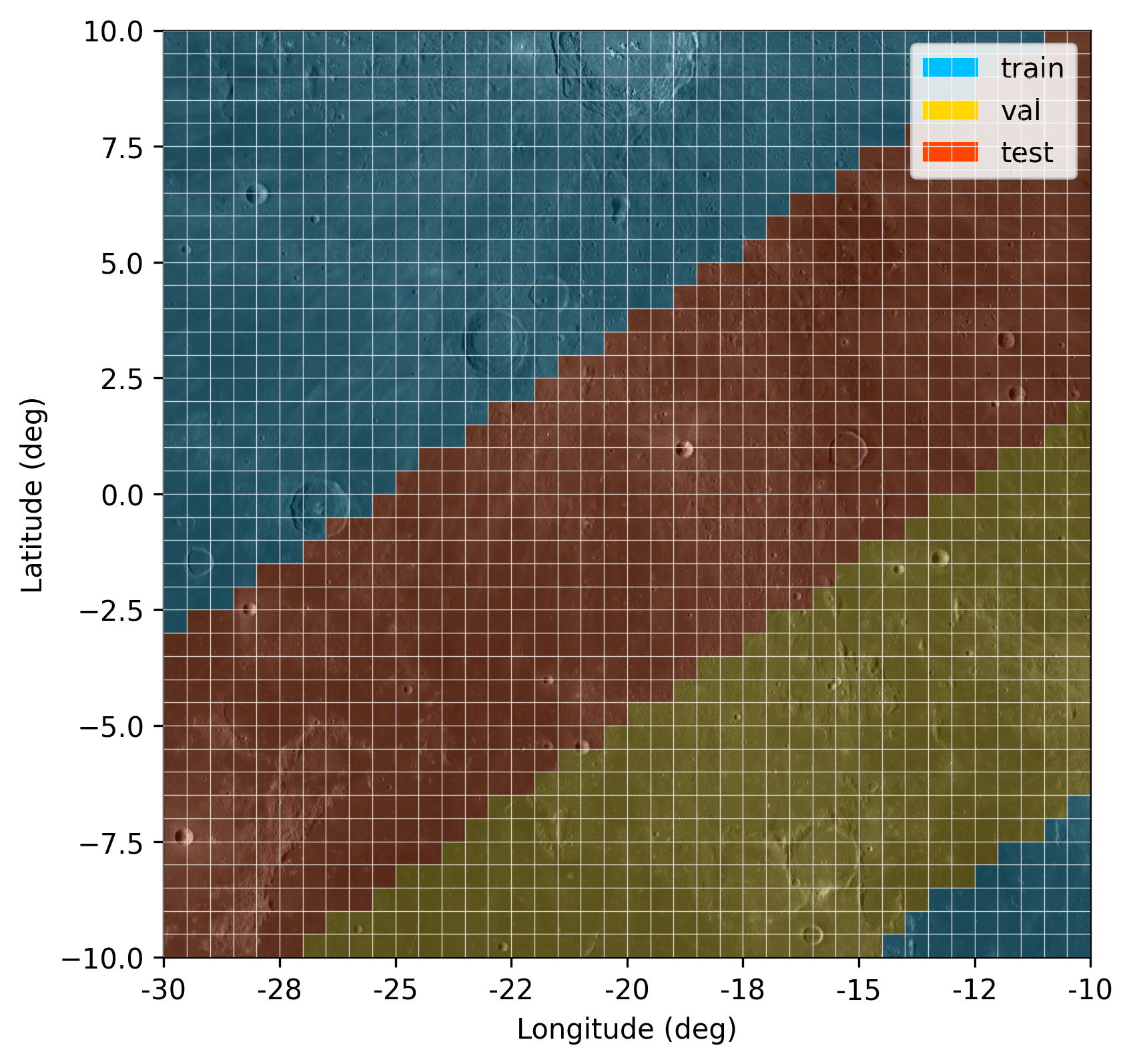}
        \caption{Zoomed view with 0.5$^\circ$ grid.}
        \label{fig:bandsplit_zoom}
    \end{subfigure}
    \caption{Visualisation of the half-degree lunar grid and the diagonal band-wise train/validation/test split. Left: global WAC background with split overlay. Right: zoomed region showing the half-degree grid over WAC with the split overlay.}
    \label{fig:grid_and_bandsplit}
\end{figure}

\end{document}